\documentclass[twocolumn,twoside]{IEEEtran}
\usepackage{amsmath,amssymb,amsthm,epsfig,dsfont,color,subfigure,empheq,graphicx,subfigure}
\usepackage{enumerate,stfloats,url,algorithm,algorithmic}
\usepackage[table]{xcolor}

\DeclareMathOperator{\rank}{rank}

\DeclareMathOperator{\sign}{sign}

\DeclareMathOperator{\trace}{tr}

\newtheorem{proposition}{Proposition}

\theoremstyle{remark}\newtheorem{remark}{Remark}

\makeatletter
\def\hlinew#1{%
  \noalign{\ifnum0=`}\fi\hrule \@height #1 \futurelet
   \reserved@a\@xhline}

\begin{document}
\bstctlcite{IEEEexample:BSTcontrol}

\title{Online Energy Price Matrix Factorization\\
for Power Grid Topology Tracking}

\author{Vassilis Kekatos,~\IEEEmembership{Member,~IEEE,} Georgios B.
Giannakis,~\IEEEmembership{Fellow,~IEEE,}  and Ross Baldick,~\IEEEmembership{Fellow,~IEEE} 
\thanks{Work in this paper was supported by the Inst. of Renewable Energy and the Environment (IREE) under grant no. RL-0010-13, University of Minnesota, and NSF grants CCF-1423316 and CCF-1442686. V. Kekatos and G. B. Giannakis are with the ECE Dept., University of Minnesota, Minneapolis, MN 55455, USA. R. Baldick is with the ECE Dept., University of Texas at Austin, TX 78712. Emails:\{kekatos,georgios\}@umn.edu, baldick@ece.utexas.edu}}

\maketitle

\begin{abstract}
Grid security and open markets are two major smart grid goals. Transparency of market data facilitates a competitive and efficient energy environment, yet it may also reveal critical physical system information. Recovering the grid topology based solely on publicly available market data is explored here. Real-time energy prices are calculated as the Lagrange multipliers of network-constrained economic dispatch; that is, via a linear program (LP) typically solved every 5 minutes. Granted the grid Laplacian is a parameter of this LP, one could infer such a topology-revealing matrix upon observing successive LP dual outcomes. The matrix of spatio-temporal prices is first shown to factor as the product of the inverse Laplacian times a sparse matrix. Leveraging results from sparse matrix decompositions, topology recovery schemes with complementary strengths are subsequently formulated. Solvers scalable to high-dimensional and streaming market data are devised. Numerical validation using real load data on the IEEE 30-bus grid provide useful input for current and future market designs.
\end{abstract}


\section{Introduction}\label{sec:intro}
An independent system operator collects energy offers and bids, and dispatches power by maximizing the social welfare while meeting physical grid limitations. To guarantee competitive market operation, multiple data are communicated to market participants or are openly publicized, either in real-time or with certain delay~\cite{Ott03}. Such market data may involve energy prices, bids and offers, congestion information, demand, and renewable generation. Looking forward, the smart grid vision calls for energy markets reaching the distribution level to promote participation, accounting for increased stochasticity at a finer time resolution~\cite{DOE}. New reliable market designs are hence to be developed.

From state estimation to load prediction, inference using data has been a major grid operation component. Facing smart grid challenges and opportunities, grid analytics are now extending to price and renewable forecasting, consumer preference learning, and cyber-physical attack detection~\cite{SPM2013}. Among grid learning tasks, topology monitoring is critical for security, market clearing, and billing. Although operators monitor grid topology via the generalized state estimator~\cite{AburExpositoBook}, topology tracking could be used for other purposes. Data attacks on the state estimator require precise physical network information~\cite{XiMo11}. Knowing congested transmission lines could assist in informed bidding or in market manipulation~\cite{LeHuBaPi11}. Line reactances could be used as a measure of electrical distance to cluster buses, or reveal influential nodes. Moreover, the Laplacian matrix of the graph representing a grid could capture the correlation across pricing nodes~\cite{KZG14}, or characterize the performance of decentralized algorithms.

Although there has been a long line of research regarding attacks on the state estimator, grid topology recovery using readily accessible market data has been overlooked. Stealth data attacks to the power system state estimator were first recognized in~\cite{LiNi11}. Their impact on state estimation and market outcomes was characterized in~\cite{KoJi11},~\cite{LangTong14}. The possibility of data framing attacks deceiving the bad data processor were explored in~\cite{framing}. Attacks and countermeasures on power system controllers have been studied in~\cite{Pasqualetti}. Procedures for detecting and identifying cyber-physical attacks have been also reported; see e.g., \cite{Sandberg}. Designing cyber-physical attacks generally presumes the grid topology to be known~\cite{XiMo11}, \cite{XuWaTang11}.

Detecting topological changes from the operator's perspective has been studied in \cite{abur10naps}, \cite{tate09pesgm}; while transmission line outages can be efficiently revealed via the sparse overcomplete representation of~\cite{ZhGi12}. Grid topology recovery is cast as a blind factorization on the matrix of spatio-temporal nodal injections in~\cite{LiPoSc13}: Even though building on the sparsity and positive semidefiniteness of the grid Laplacian, \cite{LiPoSc13} relies on linear independence across voltage phases. Considering a power line communication network, time delays of communication signals are leveraged to unveil the microgrid structure in~\cite{ErTpVi13}. By postulating a Gaussian Markov random field over nodal voltage phases, transmission network faults could also be localized~\cite{HeZhang2011}. Likewise for distribution grids, the topology recovery scheme of~\cite{BoSch13} exploits the sample covariance matrix of nodal voltage magnitudes.

All in all, existing topology recovery schemes presume access to a physical system quantity (power injections, voltage phases or magnitudes, communication delays) that is actually measured over all buses. In contrast, our previous work introduced the possibility of topology tracking using readily available cyber-system data~\cite{KeGiaBal14}. The idea is that real-time energy prices are found by the system operator as the Lagrange multipliers of the network-constrained economic dispatch, which is a linear program (LP) typically solved every 5 minutes. Dispatch decisions are the primal variables of this LP, while grid topology and electricity offers/bids are its parameters. Observing the dual variables (prices) related to multiple offer/bids instances, the crux is to recover the quasi-stationary topology underlying this LP (Section~\ref{sec:model}). Our first contribution is recognizing that properties of the Laplacian matrix and sparsity in congested lines could be exploited: The matrix of spatio-temporal prices can be factorized as the product of a doubly positive matrix with sparse inverse times a sparse matrix (Section~\ref{sec:approach}). Novel blind recovery schemes of complementary strengths constitute the second contribution (Section~\ref{sec:batch}): Different from~\cite{KeGiaBal14}, the low-rank property of one of the matrix factors is not regularized here, thus significantly simplifying the problem. As our third contribution, algorithms handling big market data are developed based on the alternating direction method of multipliers and its online version (Section~\ref{sec:online}). Advancing tools from online optimization, an algorithm handling streaming market data is devised. Distinct from~\cite{KeGiaBal14}, such an online approach is pertinent to future smart grid market designs. Experiments with market data obtained using real load data over the IEEE 30-bus benchmark corroborate the validity of our findings (Section~\ref{sec:simulations}).


\emph{Notation.} Lower- (upper-) case boldface letters denote column vectors (matrices); $\mathbf{1}$ and $\mathbf{0}$ denote the all-ones and all-zeros vectors. Symbols $\mathbf{X}'$, $\trace(\mathbf{X})$, and $|\mathbf{X}|$, stand for matrix transposition, trace, and determinant, respectively. Symbol $\mathbb{S}^N$ ($\mathbb{S}_{+}^N$) is the set of real $N\times N$ symmetric (positive semidefinite) matrices. Regarding matrix norms, $\|\mathbf{A}\|_*$ is the nuclear norm (sum of matrix singular values); $\|\mathbf{A}\|_F$ is the Frobenius norm; and $\|\mathbf{A}\|_1:=\sum_{m,n} |\mathbf{A}_{m,n}|$.

\section{Energy Price Data Model}\label{sec:model}
Before delineating our price data model, this section reviews linear DC power flows and real-time energy markets.

\subsection{Power Grid Modeling}\label{subsec:gridmodel}
Consider a power grid represented by the graph $\mathcal{G}=(\mathcal{V},\mathcal{E})$, where the set of nodes $\mathcal{V}$ corresponds to $N+1$ buses, and the edges in $\mathcal{E}$ to $L$ transmission lines. The grid topology is captured via the $L\times (N+1)$ branch-bus incidence matrix $\tilde{\mathbf{A}}$~\cite{SPM2013}. For a connected grid, the nullity of $\tilde{\mathbf{A}}$ is one; and by definition, $\tilde{\mathbf{A}}\mathbf{1} = \mathbf{0}$. If $x_l>0$ is the reactance of line $l$ and $\mathbf{D}$ an $L\times L$ diagonal matrix with $[\mathbf{D}]_{l,l}=x_l^{-1}$, the \emph{bus reactance matrix} can be defined as $\mathbf{\tilde{B}}:=\mathbf{\tilde{A}}'\mathbf{D}\mathbf{\tilde{A}}$. Given that $\mathbf{\tilde{B}}$ is the weighted Laplacian of $\mathcal{G}$, it is positive semidefinite, and $\mathbf{1}$ is an eigenvector corresponding to $\mathbf{\tilde{B}}$'s zero eigenvalue.

The DC power flow model can now be expressed in matrix-vector form. The active power flow from bus $n$ to bus $m$ over line $l$ can be approximated as $f_l=(\theta_n-\theta_m)/x_l$, where $\theta_n$ is the voltage phase at bus $n$; while the power injection at bus $n$ is $p_n=\sum_{l:(n,m)} f_l$. By stacking $\{\theta_n,p_n\}_{n=1}^{N+1}$ and $\{f_l\}_{l=1}^L$ in $\tilde{\boldsymbol{\theta}},~\tilde{\mathbf{p}}\in\mathbb{R}^{N+1}$ and $\mathbf{f}\in\mathbb{R}^L$, respectively; it follows that $\mathbf{f}=\mathbf{D}\tilde{\mathbf{A}}\tilde{\boldsymbol{\theta}}$ and $\tilde{\mathbf{p}} = \tilde{\mathbf{A}}'\mathbf{f} = \tilde{\mathbf{B}}\boldsymbol{\tilde{\theta}}$. By eliminating $\boldsymbol{\tilde{\theta}}$, the flows $\mathbf{f}$ can be linearly expressed in terms of $\mathbf{\tilde{p}}$; yet $\mathbf{\tilde{B}}$ is non-invertible. 

To resolve the singularity of $\mathbf{\tilde{B}}$, partition $\tilde{\mathbf{A}}$ into the first and the rest of its columns as $\tilde{\mathbf{A}}=[\tilde{\mathbf{a}}~\mathbf{A}]$. For a connected $\mathcal{G}$, the \emph{reduced branch-bus incidence matrix} $\mathbf{A}$ has full column-rank. Thus, the \emph{reduced bus reactance matrix} $\mathbf{B}:=\mathbf{A}'\mathbf{D}\mathbf{A}$, is strictly positive definite. Setting $\theta_1=0$, it readily follows
\begin{equation}\label{eq:f}
\mathbf{f}= \mathbf{T} \mathbf{\tilde{p}}.
\end{equation}
where $\mathbf{T}:=[\mathbf{0} ~ \mathbf{D}\mathbf{A}\mathbf{B}^{-1}]\in \mathbb{R}^{L\times (N+1)}$.

\subsection{Modeling of Real-Time Energy Markets}\label{subsec:marketmodel}
Building on this model, let us now review real-time energy markets. Energy markets determine the price for electricity by matching supply and demand. Due to time-varying demand and transmission grid limitations, the electricity cost varies across time and space (buses), giving rise to locational marginal prices (LMPs)~\cite{Ott03}. Real-time markets are spot markets where hourly power schedules determined over the previous day are adjusted every five minutes to accommodate real-time deviations. Specifically, real-time LMPs are found via the network-constrained economic dispatch, typically formulated as the following LP~\cite{ExpConCanBook} 
\begin{subequations}\label{eq:market}
\begin{align}
\mathbf{\tilde{p}}_t^*\in\arg\min_{\mathbf{\tilde{p}}_t} ~&~\mathbf{\tilde{c}}_t'\mathbf{\tilde{p}}_t\label{eq:market:cost}\\ 
\textrm{s.to}~&~\underline{\mathbf{p}}_t\leq \mathbf{\tilde{p}}_t\leq \overline{\mathbf{p}}_t\label{eq:market:box}\\
~&~ \mathbf{\tilde{p}}_t'\mathbf{1}=0\label{eq:market:balance}\\
~&~ -\overline{\mathbf{f}}\leq \mathbf{T}\mathbf{\tilde{p}}_t \leq \overline{\mathbf{f}}.\label{eq:market:flows}
\end{align}
\end{subequations}
Problem \eqref{eq:market} determines the incremental power injections $\mathbf{\tilde{p}}_t^*$ for the upcoming 5-min interval indexed by $t$. The optimum dispatch $\mathbf{\tilde{p}}_t^*$ is found by minimizing the electricity cost in \eqref{eq:market:cost}; while satisfying the power limits in \eqref{eq:market:box}, achieving the supply-demand balance via \eqref{eq:market:balance}, and confining line flows approximated by \eqref{eq:f} to lie within a secure range [cf.~\eqref{eq:market:flows}].
The power injection bounds in \eqref{eq:market:box} model bid blocks. 

By solving \eqref{eq:market}, the operator not only determines $\mathbf{\tilde{p}}_t^{\star}$, but also calculates the LMPs as follows. Let $\lambda_{0,t}$ be the optimal Lagrange multiplier associated with the supply-demand equality in \eqref{eq:market:balance}; and $(\underline{\boldsymbol{\mu}}_t,\overline{\boldsymbol{\mu}}_t)\in \mathbb{R}^{L}_{+}\times \mathbb{R}^{L}_{+}$ be the optimal Lagrange multipliers related to the lower and upper flow limits in \eqref{eq:market:flows}. By duality and upon defining $\boldsymbol{\mu}_t:=\underline{\boldsymbol{\mu}}_t-\overline{\boldsymbol{\mu}}_t$,  problem \eqref{eq:market} can be equivalently expressed as
\begin{align}\label{eq:marketLagrangian}
\mathbf{\tilde{p}}_t^*\in\arg\max_{\mathbf{\tilde{p}}_t } ~&~ 
\left( \lambda_{0,t}\mathbf{1} +\mathbf{T}'\boldsymbol{\mu}_t - \tilde{\mathbf{c}}_t\right)'
\mathbf{\tilde{p}}_t\\
\textrm{s.to}~&~\underline{\mathbf{p}}_t\leq \mathbf{\tilde{p}}_t \leq \overline{\mathbf{p}}_t.\nonumber
\end{align}
If $\lambda_{0,t}\mathbf{1} +\mathbf{T}'\boldsymbol{\mu}_t$ is selected as the vector of nodal electricity prices at time $t$ and assuming $\tilde{\mathbf{c}}_t$ are the actual marginal costs, then \eqref{eq:marketLagrangian} reveals that $\mathbf{\tilde{p}}_t^{\star}$ maximizes the sum of the individual profits. In practice, to account for transmission line losses ignored by the DC model, LMPs are calculated as
\begin{align}\label{eq:lmp}
\boldsymbol{\tilde{\pi}}_t:=\lambda_{0,t}\mathbf{1} +\left[\begin{array}{c} 0\\ \mathbf{B}^{-1}\mathbf{A}'\mathbf{D}\boldsymbol{\mu}_t\end{array}\right]  +\tilde{\boldsymbol{\ell}}_t
\end{align}
where $\tilde{\boldsymbol{\ell}}_t$ is a relatively small loss correction~\cite{ExpConCanBook}. 

The LMPs in \eqref{eq:lmp} consist of three summands: the marginal energy component (MEC) $\lambda_{0,t}$; the marginal congestion component (MCC) $[0 ~ \boldsymbol{\mu}_t' \mathbf{D}\mathbf{A} \mathbf{B}^{-1}]'$; and the marginal loss component (MLC) $\tilde{\boldsymbol{\ell}}_t$. According to \eqref{eq:marketLagrangian}, the MEC is the energy price at the reference bus (without loss of generality selected here as bus 1). When the power flow on line $l$ reaches the upper or lower limit at time $t$, that is $f_{l,t}= \overline{f}_l$ or $f_{l,t}=- \overline{f}_l$, then line $l$ is deemed \emph{congested}. Complementary slackness implies that if line $l$ is not congested at time $t$, the $l$-th entry of $\boldsymbol{\mu}_t$ is zero. Apparently, if there are no congested lines and losses were ignored, all nodes would enjoy the same energy price $\lambda_{0,t}$.

\subsection{Problem Statement}\label{subsec:problem}
Depending on the market, the three LMP components are announced either separately or collectively as a sum. In the former case, the MCC is readily available. In the latter, the effect of MEC can be isolated by subtracting the first entry of $\tilde{\boldsymbol{\pi}}_t$ from all entries of $\tilde{\boldsymbol{\pi}}_t$. It can be argued that subtracting the first entry does not harm the generality of this preprocessing step, even if the reference bus is not bus 1. Either way, collect all but the first bus prices in $\boldsymbol{\pi}_t\in\mathbb{R}^N$, for which we postulate:
\begin{align}\label{eq:lmp2}
\boldsymbol{\pi}_t &= \mathbf{B}^{-1}\mathbf{s}_t +\mathbf{n}_t
\end{align}
where $\mathbf{s}_t :=\mathbf{A}'\mathbf{D}\boldsymbol{\mu}_t$ and $\mathbf{n}_t$ captures the MLC. Slightly abusing terminology, $\boldsymbol{\pi}_t$ will be henceforth termed the LMPs.

Market clearing occurs every five minutes, and market bids $\{\mathbf{\tilde{c}}_t, \underline{\mathbf{p}}_t, \overline{\mathbf{p}}_t\}$ change partially over time, adapting to demand and generation fluctuations. Consider collecting the LMPs of \eqref{eq:lmp2} over the horizon $\mathcal{T}:=\{t:~t=1,\ldots,T\}$ of $T$ consecutive market intervals, and suppose grid topology remains invariant over $\mathcal{T}$. Upon stacking $\{\boldsymbol{\pi}_t,\mathbf{s}_t,\mathbf{n}_t\}_{t\in\mathcal{T}}$ as columns of the $N\times T$ matrices $\mathbf{\Pi}$, $\mathbf{S}$, and $\mathbf{N}$, respectively, it follows from \eqref{eq:lmp2}:
\begin{align}\label{eq:model}
\mathbf{\Pi} = \mathbf{B}^{-1}\mathbf{S} +\mathbf{N}.
\end{align}
Model \eqref{eq:model} asserts that if $\mathbf{N}$ is ignored, the price matrix $\mathbf{\Pi}$ can be factorized as the product of the inverse grid Laplacian $\mathbf{B}^{-1}$ times matrix $\mathbf{S}$. With \eqref{eq:model}, topology recovery can be now formulated as the problem of finding $(\mathbf{B},\mathbf{S})$ given $\{\boldsymbol{\pi}_t\}_{t\in\mathcal{T}}$.

\begin{remark} Having multi-block offers and several bidders per bus does not harm generality of \eqref{eq:lmp}-\eqref{eq:model}. Specifically, electricity offers and bids oftentimes consist of multiple blocks: For example, a generator may offer to produce the first 20MWh for at least 20$\$/$MWh and the next 5MWh for at least 23$\$/$MWh at the same bus. In this case, the corresponding $p_{n,t}$ should be decomposed as the sum of two extra optimization variables as $p_{n,t}=p_{n,t}^1 + p_{n,t}^2$ with $0\leq p_{n,t}^1\leq 20$ and $20\leq p_{n,t}^2\leq 25$; while the summand $c_{n,t}p_{n,t}$ in \eqref{eq:market} is replaced by $20 p_{n,t}^1+23p_{n,t}^2$. Having multiple generators and/or consumers at the same bus is handled similarly. Either way, constraints \eqref{eq:market:balance}-\eqref{eq:market:flows} apparently remain unaltered. Hence, even though simplifying, \eqref{eq:market} is sufficiently representative. Actually, Section~\ref{sec:simulations} involves tests with multi-block offers.
\end{remark}

\section{Topology Recovery Approaches}\label{sec:approach}
If the MCCs are announced separately, matrix $\mathbf{\Pi}$ satisfies the noiseless counterpart of \eqref{eq:model}, namely
\begin{align}\label{eq:nmodel}
\mathbf{\Pi} = \mathbf{B}^{-1}\mathbf{S}.
\end{align}
Decomposing $\mathbf{\Pi}$ into $(\mathbf{B},\mathbf{S})$ constitutes a blind matrix factorization problem. To uniquely recover the factors, their rich structure delineated next should be properly exploited.

Recall that $\mathbf{B}$ is positive definite. Once $\mathbf{B}$ has been recovered, the original grid Laplacian $\mathbf{\tilde{B}}$ can be trivially found in light of the property $\mathbf{\tilde{B}} \mathbf{1}= \mathbf{0}$. Note further that the $(n,m)$-th entry of $\mathbf{B}$ equals $-x_{nm}^{-1}$, if there is a line between buses $m$ and $n$; and zero otherwise. Granted power grids are sparingly connected, $\mathbf{B}$ is not only \emph{sparse}, but its off-diagonal entries are non-positive. Having positive eigenvalues and non-positive off-diagonal entries implies $\mathbf{B}$ is an invertible M-matrix~\cite[Sec.~2.5]{HornJohnson}. Hence, $\mathbf{B}^{-1}$ has positive entries, i.e., $\mathbf{B}^{-1}> \mathbf{0}$.

As far as $\mathbf{S}$ is concerned, its columns can be expressed as $\mathbf{s}_t = \sum_{l\in \mathcal{E}} \mu_{t,l}x_l^{-1} \mathbf{a}_l$. Since many of the $\{\mu_{t,l}\}_{l}$ in \eqref{eq:st} are expected to be zero [cf.~Prop.~\ref{pro:sparsemu}], $\mathbf{s}_t$ can be also written as
\begin{align}\label{eq:st}
\mathbf{s}_t = \sum_{l\in \mathcal{C}_t}  \frac{\mu_{t,l}}{x_l} \mathbf{a}_l 
\end{align}
where $\mathcal{C}_t\subseteq \mathcal{E}$ is the subset of congested lines at time $t$. In other words, $\mathbf{s}_t$ is a linear combination of few $\mathbf{a}_l$'s. Given that $\mathbf{a}_l$ are sparse, matrix $\mathbf{S}$ is expected to be sparse too. Typically, only a few transmission lines become congested over a short market period (say one day): In the California ISO (CAISO) for example, only two transmission lines are typically congested~\cite{Price11}. Hence, it can be assumed that the $\{\mathcal{C}_t\}_{t=1}^T$ overlap significantly, or that the locations of the non-zero entries of $\{\boldsymbol{\mu}_t\}_{t=1}^T$ remain relatively time-invariant. Since $\mathbf{s}_t$'s are linear combinations of those few $\mathbf{a}_l$'s corresponding to congested lines, $\mathbf{S}$ is expected to exhibit low rank. The invertibility of $\mathbf{B}$ implies $\mathbf{\Pi}$ should have low rank too.

It will be useful also to recognize that the factorization in \eqref{eq:nmodel} is scale-invariant: If $(\mathbf{B},\mathbf{S})$ satisfies \eqref{eq:nmodel}, so does the pair $(\alpha\mathbf{B},\alpha\mathbf{S})$ for all $\alpha>0$. To waive this inherent ambiguity, the maximum diagonal entry of $\mathbf{B}$ is assumed to be unity. Due to this normalization, matrix $\mathbf{B}$ should satisfy $\mathbf{B}\succeq \mathbf{0}$ and $\mathbf{B}\leq \mathbf{I}$. 

Leveraging these properties, one could recover $(\mathbf{B},\mathbf{S})$ by solving the optimization problem:
\begin{subequations}\label{eq:ell0}
\begin{align}
\min_{\mathbf{B},\mathbf{S}} ~&~ \|\mathbf{S}\|_0 + \kappa_0 \|\mathbf{B}\|_0  \label{eq:ell0:cost}\\ 
\textrm{s.to}~&~\mathbf{B\Pi}=\mathbf{S},~\mathbf{B}\succ \mathbf{0},~\mathbf{B}\leq \mathbf{I}\label{eq:ell0:constraints}
\end{align}
\end{subequations}
where $\|\mathbf{X}\|_0$ is the $\ell_0$-(pseudo)norm of a matrix counting its non-zero entries, and $\kappa_0>0$ is a parameter balancing the sparsity between the two matrices. Problem \eqref{eq:ell0} finds the sparsest pair $(\mathbf{B},\mathbf{S})$ that satisfies model \eqref{eq:nmodel} and the structural constraints of $\mathbf{B}$. Different from \cite{KeGiaBal14}, the rank of $\mathbf{S}$ is not penalized here, since $\mathbf{B\Pi}=\mathbf{S}$ and the invertibility of $\mathbf{B}$ enforce $\rank(\mathbf{S})=\rank(\mathbf{\Pi})$ anyway.

Minimizing the $\ell_0$-norm is in general NP-hard~\cite{Natarajan}. Following advances in compressive sensing~\cite{Donoho}, the $\ell_0$-norm will be surrogated by the $\ell_1$-norm to yield the convex problem
\begin{align}\label{eq:ell1}
\min_{\mathbf{B},\mathbf{S}} ~&~ \|\mathbf{S}\|_1 + \kappa_1 \trace ( \mathbf{P}\mathbf{B})-\kappa_2 \log|\mathbf{B}|\\ 
\textrm{s.to}~&~\mathbf{B\Pi}=\mathbf{S},~ \mathbf{B}\in\mathcal{B}.\nonumber
\end{align}
where $\mathbf{P}:=\mathbf{I}-\mathbf{1}\mathbf{1}'$, $\mathcal{B}:=\left\{\mathbf{B}:\mathbf{B}\succeq \mathbf{0},~\mathbf{B}\leq \mathbf{I}\right\}$, and $\kappa_1,\kappa_2>0$. Two observations are in order regarding \eqref{eq:ell1}.

First, since that the diagonal entries of $\mathbf{B}$ are strictly positive, $\|\mathbf{B}\|_0$ in \eqref{eq:ell0:cost} has been replaced by the sum of the off-diagonal entries of $\mathbf{B}$ in their absolute values, that is
\begin{align*}
\sum_{n,m:n\neq m}|\mathbf{B}_{n,m}| 
&= \sum_{n} \mathbf{B}_{n,n} - \sum_{n,m}\mathbf{B}_{n,m}\\
&=\trace(\mathbf{B}) -\mathbf{1}'\mathbf{B}\mathbf{1}=\trace(\mathbf{P}\mathbf{B})
\end{align*}
where the first equality comes from the non-positive off-diagonal entries of $\mathbf{B}$, and the rest from properties of the trace. 

Second, ideally $\mathbf{B}$ should be enforced to be strictly positive definite, i.e., $\mathbf{B}\succ \mathbf{0}$. Nonetheless, current optimization algorithms cannot guarantee the minimizer to lie in the interior of the feasible set. On the other hand, imposing $\mathbf{B}\succeq \mathbf{0}$ admits the trivial solution $(\mathbf{B},\mathbf{S})=(\mathbf{0},\mathbf{0})$. As a remedy, the term $-\log|\mathbf{B}|$ has been added in the cost of \eqref{eq:ell1} to confine $\mathbf{B}$ in the interior of the positive semidefinite cone $\mathbf{B}\succeq \mathbf{0}$.

By eliminating $\mathbf{S}$, \eqref{eq:ell1} can be equivalently transformed to
\begin{align}\label{eq:ell1B}
\min_{\mathbf{B}\in\mathcal{B}} ~&~ \|\mathbf{B}\mathbf{\Pi}\|_1 + \kappa_1 \trace ( \mathbf{P}\mathbf{B})-\kappa_2 \log|\mathbf{B}|.
\end{align}
The strict convexity of $-\kappa_2\log|\mathbf{B}|$ guarantees that \eqref{eq:ell1B} and hence \eqref{eq:ell1} have unique minimizers.

When $\{\boldsymbol{\pi}_t\}_{t\in\mathcal{T}}$ comprise of both MEC and MLC, model \eqref{eq:model} is more pertinent than the exact model in \eqref{eq:nmodel}. The non-convex problem in \eqref{eq:ell0} could be then replaced by
\begin{align}\label{eq:ell02}
\min_{\mathbf{B}\in\mathcal{B},\mathbf{S}} ~&~\tfrac{1}{2}\|\mathbf{B}\mathbf{\Pi}-\mathbf{S}\|_F^2 + \kappa_{1} \|\mathbf{B}\|_0\\
~&~ +\kappa_{2}\|\mathbf{S}\|_0 +\kappa_{3}\rank(\mathbf{S})\nonumber
\end{align}
for $\kappa_{1},\kappa_{2},\kappa_{3}>0$. The approach in \eqref{eq:ell02} aims at minimizing the least-squares distance between $\mathbf{B\Pi}$ and $\mathbf{S}$, while seeking sparse $(\mathbf{B},\mathbf{S})$ and a low-rank $\mathbf{S}$. However, minimizing the $\ell_0$-norm and the matrix rank constitutes an NP-hard problem. In the same spirit \eqref{eq:ell0} was surrogated by \eqref{eq:ell1}, the hard problem in \eqref{eq:ell02} is approximated by the convex problem
\begin{align}\label{eq:ell12}
\min_{\mathbf{B}\in\mathcal{B},\mathbf{S}} ~&~\tfrac{1}{2}\|\mathbf{B\Pi}-\mathbf{S}\|_F^2 + \kappa_{1} \|\mathbf{B}\|_1 +\kappa_{2}\|\mathbf{S}\|_1 \\
&+\kappa_{3}\|\mathbf{S}\|_*-\kappa_4\log |\mathbf{B}|\nonumber
\end{align}
where $\{\kappa_i>0\}_{i=1}^4$ and $\|\mathbf{S}\|_*$ serves as a convex surrogate for $\rank(\mathbf{S})$. A solver and recovery results from \eqref{eq:ell12} can be found in \cite{KeGiaBal14}. Given that MCCs are typically announced independently, our focus will be henceforth on model \eqref{eq:model}.


\section{Batch Topology Recovery Scheme}\label{sec:batch}
Although \eqref{eq:ell1}-\eqref{eq:ell1B} could be solved by commercial software for relatively small problems, interior point-based solvers cannot handle $N$ and $T$ larger than a few hundreds. There are two main optimization challenges here: the objective term $\|\mathbf{B\Pi}\|_1$ and the feasible set $\mathcal{B}$. Regarding the former, not only it is non-differentiable, but also involves a linear transformation of $\mathbf{B}$. Note that $\mathcal{B}$ is the intersection of the positive definite cone and a shifted version of the positive cone. Albeit projection over each of these cones is relatively easy, there is no closed-form solution for projecting on $\mathcal{B}$.

Given these challenges, an algorithm based on the alternating direction method of multipliers (ADMM) is derived next. ADMM targets solving problems of the form~\cite{oADMM}
\begin{align}\label{eq:ADMM}
\min_{\mathbf{x}\in\mathcal{X},\mathbf{z}\in\mathcal{Z}} & \left\{f(\mathbf{x}) + g(\mathbf{z}):~\mathbf{F}\mathbf{x} + \mathbf{G}\mathbf{z}=\mathbf{c}\right\}
\end{align}
where $f(\mathbf{x})$ and $g(\mathbf{z})$ are convex functions; $\mathcal{X}$ and $\mathcal{Z}$ are convex sets; and $(\mathbf{F},\mathbf{G},\mathbf{c})$ model the linear equality constraints coupling variables $\mathbf{x}$ and $\mathbf{z}$. In its normalized form, ADMM assigns a Lagrange multiplier $\mathbf{y}$ for the equality constraint and solves \eqref{eq:ADMM} by iterating over the recursions
\begin{subequations}\label{eq:ADMM:steps}
\begin{align}
\mathbf{x}^{i+1}&:=\arg\min_{\mathbf{x}\in\mathcal{X}} ~f(\mathbf{x}) + \tfrac{\rho}{2}\|\mathbf{F}\mathbf{x} + \mathbf{G}\mathbf{z}^i-\mathbf{c} + \mathbf{y}^i\|_2^2\label{eq:ADMM:S1}\\
\mathbf{z}^{i+1}&:=\arg\min_{\mathbf{z}\in\mathcal{Z}} ~g(\mathbf{z}) + \tfrac{\rho}{2}\|\mathbf{F}\mathbf{x}^{i+1}  + \mathbf{G}\mathbf{z}-\mathbf{c} + \mathbf{y}^i\|_2^2\label{eq:ADMM:S2}\\
\mathbf{y}^{i+1}&:=\mathbf{y}^{i}  + \mathbf{F}\mathbf{x}^{i+1} + \mathbf{G}\mathbf{z}^{i+1}-\mathbf{c}.\label{eq:ADMM:S3}
\end{align}
\end{subequations} 
 for some $\rho>0$. To apply ADMM and end up in efficient updates for \eqref{eq:ell1}, we first replace variable $\mathbf{B}$ with three copies $\mathbf{B}_1$, $\mathbf{B}_2$, and $\mathbf{B}_3$, to yield the equivalent problem
\begin{subequations}\label{eq:ell1ADMM}
\begin{align}
\min_{\mathbf{B}_1, \mathbf{B}_2\leq \mathbf{I}, \mathbf{B}_3\succeq \mathbf{0}, \mathbf{S}} &~ \|\mathbf{S}\|_1 + \kappa_1 \trace ( \mathbf{P}\mathbf{B}_1)-\kappa_2 \log|\mathbf{B}_3|  \label{eq:ell1ADMM:cost}\\ 
\textrm{s.to}~&~\mathbf{B}_1=\mathbf{B}_2\label{eq:ell1ADMM:con1}\\
&~\mathbf{B}_1=\mathbf{B}_3\label{eq:ell1ADMM:con2}\\
&~\mathbf{B}_1\mathbf{\Pi}=\mathbf{S}\label{eq:ell1ADMM:con3}
\end{align}
\end{subequations}

Let $\mathbf{M}_{12}$, $\mathbf{M}_{13}$, and $\mathbf{M}$, denote the Lagrange multipliers corresponding to \eqref{eq:ell1ADMM:con1}, \eqref{eq:ell1ADMM:con2}, and \eqref{eq:ell1ADMM:con3}, respectively. Partitioning variables into $\mathbf{B}_1$ and $(\mathbf{B}_2,\mathbf{B}_3,\mathbf{S})$, ADMM iterates through the next three steps.

At the \emph{first step} of iterate $i$, the variable $\mathbf{B}_1$ is updated given $(\mathbf{B}_2^i,\mathbf{B}_3^i,\mathbf{S}^i)$ and $(\mathbf{M}_{12}^i,\mathbf{M}_{13}^i,\mathbf{M}^i)$ by solving
\begin{align}\label{eq:B1p}
\min_{\mathbf{B}_1} ~&~\kappa_1 \trace ( \mathbf{P}\mathbf{B}_1) 
+\tfrac{\rho}{2} \|\mathbf{B}_1-\mathbf{B}_2^i+\mathbf{M}_{12}^i\|_F^2\\
&~+\tfrac{\rho}{2} \|\mathbf{B}_1-\mathbf{B}_3^i+\mathbf{M}_{13}^i\|_F^2
+\tfrac{\rho}{2} \|\mathbf{B}_1\mathbf{\Pi}-\mathbf{S}^i+\mathbf{M}^i\|_F^2.\nonumber
\end{align}
The solution of \eqref{eq:B1p} is provided in closed form as $\mathbf{B}_1^{i+1} =  ( \mathbf{B}_2^i  -\mathbf{M}_{12}^i  + \mathbf{B}_{3}^i - \mathbf{M}_{13}^i  +(\mathbf{S}^i -  \mathbf{M}^i) \mathbf{\Pi}' -\tfrac{\kappa_1}{\rho}\mathbf{P}) \left( 2\mathbf{I} +\mathbf{\Pi}\mathbf{\Pi}' \right)^{-1}$.

During the \emph{second step}, ADMM updates the second block of variables $(\mathbf{B}_2,\mathbf{B}_3,\mathbf{S})$ given $\mathbf{B}_1^{i+1}$ and $(\mathbf{M}_{12}^i,\mathbf{M}_{13}^i,\mathbf{M}^i)$. Yet the optimization decouples over the three variables. Specifically, variable $\mathbf{B}_2$ is updated as the solution of 
\begin{equation}\label{eq:B2p}
\min_{\mathbf{B}_2\leq \mathbf{I}} ~\tfrac{\rho}{2}\|\mathbf{B}_1^{i+1}-\mathbf{B}_2+\mathbf{M}_{12}^i\|_F^2
\end{equation}
whose minimizer is $\mathbf{B}_2^{i+1} =\min\left\{\mathbf{B}_1^{i+1} + \mathbf{M}_{12}^i,\mathbf{I}\right\}$,  where the minimum operator is understood entry-wise.

Variable $\mathbf{B}_3$ is updated as the minimizer of
\begin{equation}\label{eq:B3p}
\min_{\mathbf{B}_3\succeq \mathbf{0}} ~\tfrac{1}{2}\|\mathbf{B}_1^{i+1}-\mathbf{B}_3+\mathbf{M}_{13}^i\|_F^2 -\tfrac{\kappa_2}{\rho}\log|\mathbf{B}_3|
\end{equation}
which can be readily found as follows~\cite[Lemma~1]{KeGiaBal14}: Define the operator $\mathcal{P}_{\alpha}:\mathbb{R}^{N\times N}\rightarrow \mathbb{S}_+^N$ for some $\alpha>0$ as
\begin{align}\label{eq:EVDoperator}
\mathcal{P}_{\alpha}\left(\mathbf{X}\right)= \tfrac{1}{2} 
\mathbf{V} \left( \mathbf{\Xi} + \left(\mathbf{\Xi}^2 + 4\alpha \mathbf{I}\right)^{1/2}  \right)\mathbf{V}'
\end{align}
where $\mathbf{V}\mathbf{\Xi}\mathbf{V}'$ is the eigenvalue decomposition of the symmetric matrix $\tfrac{1}{2}(\mathbf{X}+\mathbf{X}')$. Then, the solution to \eqref{eq:B3p} is
\begin{align}\label{eq:B3}
\mathbf{B}_3^{i+1}&= \mathcal{P}_{\kappa_2/\rho} \left( \mathbf{B}_1^{i+1}+\mathbf{M}_{13}^i\right).
\end{align}

Variable $\mathbf{S}$ is updated by solving
\begin{equation}\label{eq:Sp}
\min_{\mathbf{S}} ~\tfrac{1}{2}\|\mathbf{B}_1^{i+1}\mathbf{\Pi}-\mathbf{S}+\mathbf{M}^i\|_F^2 + \tfrac{1}{\rho} \|\mathbf{S}\|_1.
\end{equation}
Problem \eqref{eq:Sp} is separable across the entries of $\mathbf{S}$, admitting the closed-form minimizer:
\begin{align}\label{eq:S}
\mathbf{S}^{i+1}&=\mathcal{S}_{\rho^{-1}}\left(\mathbf{B}_1^{i+1}\mathbf{\Pi}+\mathbf{M}^i\right)
\end{align}
where $\mathcal{S}_{\beta}(x)$ is the soft thresholding operator defined as
\begin{align*}
\mathcal{S}_{\beta}(x):=
\left\{
\begin{array}{cc}
x-\beta\sign(x), & |x|> \beta\\
0, & |x|\leq \beta\\
\end{array}
\right.
\end{align*}
applied entry-wise in \eqref{eq:S}.

In the \emph{third step}, the Lagrange multipliers are updated as
\begin{align}\label{eq:Lagrange}
\mathbf{M}_{12}^{i+1} & = \mathbf{M}_{12}^{i} + \mathbf{B}_1^{i+1} - \mathbf{B}_2^{i+1}\\
\mathbf{M}_{13}^{i+1} & = \mathbf{M}_{13}^{i} + \mathbf{B}_1^{i+1} - \mathbf{B}_3^{i+1}\nonumber\\
\mathbf{M}^{i+1} & = \mathbf{M}^{i} + \mathbf{B}_1^{i+1}\mathbf{\Pi} - \mathbf{S}^{i+1}.\nonumber
\end{align}

\begin{algorithm}[t]
\caption{Batch Topology Recovery Scheme} \label{alg:batch}
\begin{algorithmic}[1]
\REQUIRE Price matrix $\mathbf{\Pi}$ and $(\kappa_1,\kappa_2,\rho)$.
\STATE Initialize $\mathbf{B}_1^0=\mathbf{B}_2^0=\mathbf{B}_3^0=\mathbf{I}_N$ and $\mathbf{S}^0=\mathbf{\Pi}$.
\STATE Initialize $\mathbf{M}_{12}^0= \mathbf{M}_{13}^0=\mathbf{0}_{N}$ and  $\mathbf{M}^0=\mathbf{0}_{N\times T}$.
\FOR{$i=1,2,\ldots,$} 
\STATE Update $\mathbf{B}_1^{i+1}$ from \eqref{eq:B1p}.
\STATE Update $\mathbf{B}_2^{i+1}$ from \eqref{eq:B2p}.
\STATE Update $\mathbf{B}_3^{i+1}$ from \eqref{eq:B3}.
\STATE Update $\mathbf{S}^{i+1}$ from \eqref{eq:S}.
\STATE Update multipliers $(\mathbf{M}_{12}^{i+1},\mathbf{M}_{13}^{i+1},\mathbf{M}^{i+1})$ from \eqref{eq:Lagrange}.
\ENDFOR
\end{algorithmic}
\end{algorithm}

The algorithm is summarized as Alg.~\ref{alg:batch}, and its convergence is guaranteed for all $\rho>0$~\cite{oADMM}.

\section{Grid Topology Tracking}\label{sec:online}
The topology recovery scheme of Section~\ref{sec:batch} presumes that:\\
\hspace*{1em} \textbf{(c1)} the power system topology remains unchanged, and\\
\hspace*{1em} \textbf{(c2)} prices are available for the entire period $\mathcal{T}$.\\
In reality, future power grids may be reconfigured frequently for dispatching and maintenance, while real-time LMPs are expected to be announced at a fast rate over thousands of buses; hence, rendering conditions (c1)-(c2) unrealistic.

To cope with these challenges, we first modify the recovery scheme of \eqref{eq:ell1} to address (c1). Specifically, rather than enforcing the constraint $\mathbf{B}\mathbf{\Pi}=\mathbf{S}$, our idea here is to look for sparse $\mathbf{S}$ yielding a small least-squares error $\|\mathbf{B\Pi}-\mathbf{S}\|_F^2$ by solving 
\begin{align*}
\min_{\mathbf{B}\in\mathcal{B},\mathbf{S}} ~&~ \tfrac{1}{2}\|\mathbf{B}\mathbf{\Pi}-\mathbf{S}\|_F^2 +\kappa_3 \|\mathbf{S}\|_1 + \kappa_1 \trace ( \mathbf{P}\mathbf{B})-\kappa_2 \log|\mathbf{B}|
\end{align*}
for $\kappa_3>0$. Upon eliminating $\mathbf{S}$, the last minimization can be shown to be equivalent to
\begin{align}\label{eq:Huber}
\min_{\mathbf{B}\in\mathcal{B}} ~&~ \tilde{h}_{\kappa_3}(\mathbf{B}\mathbf{\Pi}) + \kappa_1 \trace ( \mathbf{P}\mathbf{B})-\kappa_2 \log|\mathbf{B}|
\end{align}
where $\tilde{h}_{\kappa_3}(\mathbf{X}):=\sum_{m,n} h_{\kappa_3}(\mathbf{X}_{m,n})$, and $h_{\kappa_3}$ is the so termed \emph{Huber function}
\begin{align}\label{eq:huber}
h_{\kappa_3}(x):=\left\{\begin{array}{ll}
\tfrac{1}{2} x^2 &, |x|\leq \kappa_3\\
\kappa_3|x| - \tfrac{\kappa_3^2}{2} &, |x|> \kappa_3
\end{array}\right..
\end{align}
Notice that in \eqref{eq:ell1B}, the entries of $\mathbf{B}\mathbf{\Pi}$ are arguments of the absolute value cost. In contrast, the objective in \eqref{eq:Huber} penalizes small entries of $\mathbf{B\Pi}$ with a quadratic cost, and large ones with the absolute value cost.

To cope with (c2), solvers for \emph{streaming} rather than batch pricing data are developed next. The desiderata here is online schemes where topology estimates $\mathbf{B}^t$ are updated every time a price vector $\boldsymbol{\pi}_t$ is publicized. Advances from online optimization are particularly suitable for this task~\cite{COMID}. Tailored to big data processing, many online convex optimization algorithms aim at solving problems of the form
\begin{equation}\label{eq:OCO}
\min_{\mathbf{x}\in\mathcal{X}}~\sum_{t=1}^T \left(f_t(\mathbf{x}) + g(\mathbf{x})\right)
\end{equation}
where $f_t(\mathbf{x})$ depends on the $t$-th datum, and $g(\mathbf{x})$ is a regularizer, i.e., a function leveraging prior information on $\mathbf{x}$. 

%

Tailoring our grid topology recovery task to the online optimization setup, consider the general problem
\begin{align}\label{eq:online}
\min_{\mathbf{B}\in\mathcal{B}} ~&~ \sum_{t=1}^T\left( f_{\boldsymbol{\pi}_t}(\mathbf{B}) + \tfrac{\kappa_1}{T} \trace ( \mathbf{P}\mathbf{B})-\tfrac{\kappa_2}{T} \log|\mathbf{B}|\right).
\end{align}
By selecting $f_{\boldsymbol{\pi}_t}(\mathbf{B}):=\|\mathbf{B}\boldsymbol{\pi}_t\|_1$, problem \eqref{eq:online} yields \eqref{eq:ell1B}; whereas, for $f_{\boldsymbol{\pi}_t}(\mathbf{B}):=\tilde{h}_{\kappa_3}(\mathbf{B}\boldsymbol{\pi}_t)$, \eqref{eq:online} is equivalent to \eqref{eq:Huber}. Apparently, $f_{\boldsymbol{\pi}_t}(\mathbf{B})$ is price-dependent, and the other two terms in the objective of \eqref{eq:online} can be thought of as regularizers for $\mathbf{B}$. To solve \eqref{eq:online}, we resorted to the online ADMM algorithm of~\cite{oADMM} that cycles through:
\begin{subequations}\label{eq:oADMM:steps}
\begin{align}
\mathbf{x}^{t+1}&:=\arg\min_{\mathbf{x}\in\mathcal{X}} ~f_t(\mathbf{x}) + \tfrac{\rho}{2}\|\mathbf{F}\mathbf{x} + \mathbf{G}\mathbf{z}^t-\mathbf{c} + \mathbf{y}^t\|_2^2\label{eq:oADMM:S1} \\ 
&\qquad \qquad \quad+\tfrac{\eta}{2} \|\mathbf{x}-\mathbf{x}^t\|_2^2\nonumber\\
\mathbf{z}^{t+1}&:=\arg\min_{\mathbf{z}\in\mathcal{Z}} ~g(\mathbf{z}) + \tfrac{\rho}{2}\|\mathbf{F}\mathbf{x}^{t+1}  + \mathbf{G}\mathbf{z}-\mathbf{c} + \mathbf{y}^t\|_2^2\label{eq:oADMM:S2}\\
\mathbf{y}^{t+1}&:=\mathbf{y}^{t}  + (\mathbf{F}\mathbf{x}^{t+1} + \mathbf{G}\mathbf{z}^{t+1}-\mathbf{c}).\label{eq:oADMM:S3}
\end{align}
\end{subequations} 
Comparing \eqref{eq:oADMM:steps} with its batch counterpart in \eqref{eq:ADMM:steps}, the iteration index $i$ in \eqref{eq:oADMM:steps} coincides with the data index $t$, while the first step in \eqref{eq:oADMM:S1} entails only the current $f_t(\mathbf{x})$ together with the proximal term $\tfrac{\eta}{2} \|\mathbf{x}-\mathbf{x}^t\|_2^2$ for some $\eta>0$. Regarding its convergence, the algorithm attains sublinear regret in terms of both the cost and the constraint violation~\cite[Th.~4]{oADMM}.

Building on \eqref{eq:ell1ADMM}, introduce copies of $\mathbf{B}$ to express \eqref{eq:online} as
 \begin{subequations}\label{eq:ell1online}
\begin{align}
\min_{\mathbf{B}_1, \mathbf{B}_2, \mathbf{B}_3} & \sum_{t=1}^T\left(f_{\boldsymbol{\pi}_t}(\mathbf{B}_1) + \tfrac{\kappa_1}{T} \trace ( \mathbf{P}\mathbf{B}_1)- \tfrac{\kappa_2}{T} \log|\mathbf{B}_3| \right) \label{eq:ell1online:cost}\\ 
\textrm{s.to}~&~\mathbf{B}_1=\mathbf{B}_2\label{eq:ell1online:con1}\\
~&~\mathbf{B}_1=\mathbf{B}_3\label{eq:ell1online:con2}\\
~&~\mathbf{B}_3\succeq \mathbf{0},~\mathbf{B}_2\leq \mathbf{I}.\label{eq:ell1online:con3}
\end{align}
\end{subequations}
Similarly to \eqref{eq:ell1ADMM}, let $\mathbf{M}_{12}$ and $\mathbf{M}_{13}$ be the Lagrange multipliers corresponding to constraints \eqref{eq:ell1online:con1} and \eqref{eq:ell1online:con2}, respectively.

As soon as the $t$-th price vector $\boldsymbol{\pi}_t$ is announced, a cycle of the online ADMM of \eqref{eq:oADMM:steps} is initiated. In its \emph{first step}, $\mathbf{B}_1$ is updated via \eqref{eq:oADMM:S1}, which upon completing the squares yields
\begin{align}\label{eq:oADMMs1}
\mathbf{B}_1^t:=\arg\min_{\mathbf{B}_1} ~\tfrac{1}{2\rho+\eta}f_{\boldsymbol{\pi}_t}(\mathbf{B}_1) 
+ \tfrac{1}{2} \|\mathbf{B}_1-\check{\mathbf{B}}_1^{t-1}\|_F^2
\end{align}
where $\check{\mathbf{B}}_1^{t-1}:=\tfrac{\rho}{2\rho+\eta}    (\mathbf{B}_2^{t-1} + \mathbf{B}_3^{t-1}-\mathbf{M}_{12}^{t-1} - \mathbf{M}_{13}^{t-1}) +\tfrac{\eta}{2\rho+\eta} \mathbf{B}_1^{t-1} - \tfrac{\kappa_1}{2T(2\rho+\eta)}\mathbf{P}$. Problem \eqref{eq:oADMMs1} could be reformulated and solved as a linearly-constrained quadratic program. Interestingly, the minimizer of \eqref{eq:oADMMs1} can be found in closed form for both choices of $f_{\boldsymbol{\pi}_t}(\mathbf{B}_1)$. Specifically, if $f_{\boldsymbol{\pi}_t}(\mathbf{B}_1)=\|\mathbf{B}_1\boldsymbol{\pi}_t\|_1$, the next claim is shown in the Appendix:

\begin{proposition}\label{pro:ell1}
Given $(\mathbf{Y},\mathbf{z})\in \mathbb{R}^{M{\times} N}{\times} \mathbb{R}^N$, the minimizer
\begin{align}\label{eq:Pell1}
\hat{\mathbf{X}}:=\arg\min_{\mathbf{X}} ~ \|\mathbf{X}\mathbf{z}\|_1 + \tfrac{1}{2}\|\mathbf{X}-\mathbf{Y}\|_F^2
\end{align}
is given by
$\hat{\mathbf{X}}=\mathbf{Y}-\mathcal{S}_z\left(\mathbf{Y}\mathbf{z}\right)\mathbf{z}'$,
where $z:=\|\mathbf{z}\|_2^2$, and the operator $\mathcal{S}_z\left(\mathbf{x}\right):\mathbb{R}^N\rightarrow \mathbb{R}^N$ is defined as $\mathcal{S}_z(x):=\sign(x)\cdot \min\{\tfrac{|x|}{z},1\}$ applied entry-wise.
\end{proposition}

By Prop.~\ref{pro:ell1}, $\mathbf{B}_1^t $ can be found as a rank-one update of $\check{\mathbf{B}}_1^{t-1}$
\begin{align}\label{eq:B1online}
\mathbf{B}_1^t = \check{\mathbf{B}}_1^{t-1} - \mathcal{S}_{\|\check{\boldsymbol{\pi}}_t\|_2^2}\left(\check{\mathbf{B}}_1^{t-1}\check{\boldsymbol{\pi}}_t\right) \check{\boldsymbol{\pi}}_t'
\end{align}
where $\check{\boldsymbol{\pi}}_t:=\tfrac{1}{2\rho+\eta}\boldsymbol{\pi}_t$. The key observation here is that having a single $\ell_1$-norm $\|\mathbf{B}_1\boldsymbol{\pi}_t\|_1$ rather than $\|\mathbf{B}_1\mathbf{\Pi}\|_1=\sum_{t=1}^T\|\mathbf{B}_1\boldsymbol{\pi}_t\|_1$ [cf.~\eqref{eq:ell1B}] enabled the simple update of \eqref{eq:B1online}.

For the Huber cost, the next claim is shown in the Appendix:
\begin{proposition}\label{pro:huber}
Given $(\mathbf{Y},\mathbf{z})\in \mathbb{R}^{M{\times} N}{\times} \mathbb{R}^N$, the minimizer
\begin{align}\label{eq:Phuber}
\hat{\mathbf{X}}:=\arg\min_{\mathbf{X}} ~ \alpha\tilde{h}_{\kappa}\left(\mathbf{X}\mathbf{z}\right) + \tfrac{1}{2}\|\mathbf{X}-\mathbf{Y}\|_F^2
\end{align}
for $\alpha>0$ is given by
$\hat{\mathbf{X}}=\mathbf{Y}-\mathcal{H}_{z,\alpha,\kappa}\left(\mathbf{Y}\mathbf{z}\right)\mathbf{z}'$,
where $z:=\|\mathbf{z}\|_2^2$, and $\mathcal{H}_{z,\alpha,\kappa}\left(\mathbf{x}\right):\mathbb{R}^N\rightarrow \mathbb{R}^N$ is defined as
\begin{equation}\label{eq:Harg}
\mathcal{H}_{z,\alpha,\kappa}(x):=\left\{\begin{array}{ll}
\frac{x}{\alpha^{-1}+z} &,~ |x|\leq \kappa(1+\alpha z)\\
\alpha\kappa &, x>\kappa (1+\alpha z)\\
-\alpha\kappa &, x<-\kappa (1+\alpha z)  
\end{array}\right.
\end{equation} 
applied entry-wise.
\end{proposition}

Based on Proposition~\ref{pro:huber}, when $f_{\boldsymbol{\pi}_t}(\mathbf{B}_1)=\tilde{h}_{\kappa_3}(\mathbf{B}_1\boldsymbol{\pi}_t)$, the minimizer of \eqref{eq:oADMMs1} is
\begin{align}\label{eq:B1onlinehuber}
\mathbf{B}_1^t = \check{\mathbf{B}}_1^{t-1} - \mathcal{H}_{\|\boldsymbol{\pi}_t\|_2^2,(2\rho+\eta)^{-1},\kappa_3}\left(\check{\mathbf{B}}_1^{t-1}\boldsymbol{\pi}_t\right) \boldsymbol{\pi}_t'.
\end{align}

During the \emph{second step} of iteration $t$, matrices $(\mathbf{B}_2,\mathbf{B}_3)$ are updated similarly to \eqref{eq:B2p}-\eqref{eq:B3} as
\begin{align}
\mathbf{B}_2^{t+1}&=\min\left\{\mathbf{B}_1^{t+1}+\mathbf{M}_{12}^t,\mathbf{I}\right\}\label{eq:B2online}\\
\mathbf{B}_3^{t+1}&= \mathcal{P}_{\kappa_2/(T\rho)} \left(
\mathbf{B}_1^{t+1}+\mathbf{M}_{13}^t\right).\label{eq:B3online}
\end{align}

At the \emph{third step}, the Lagrange multipliers are updated as
\begin{align}
\mathbf{M}_{12}^{t+1} & = \mathbf{M}_{12}^{t} + \mathbf{B}_1^{t+1} - \mathbf{B}_2^{t+1}\label{eq:LM12online}\\
\mathbf{M}_{13}^{t+1} & = \mathbf{M}_{13}^{t} + \mathbf{B}_1^{t+1} - \mathbf{B}_3^{t+1}.\label{eq:LM13online}
\end{align}
To summarize, grid topology recovery using inexact streaming pricing data can be performed using Algorithm~\ref{alg:online}.

\begin{algorithm}[t]
\caption{Topology Recovery Tracking Scheme} \label{alg:online}
\begin{algorithmic}[1]
\REQUIRE $(\kappa_1,\kappa_2,\kappa_3,\rho,\eta)$.
\STATE Initialize $\mathbf{B}_1^0=\mathbf{B}_2^0=\mathbf{B}_3^0=\mathbf{I}_N$.
\STATE Initialize $\mathbf{M}_{12}^0= \mathbf{M}_{13}^0=\mathbf{0}_{N}$.
\FOR{$t=1,2,\ldots,T$} 
\STATE Acquire price vector $\boldsymbol{\pi}_t$.
\STATE Update $\mathbf{B}_1^{t+1}$ using \eqref{eq:B1online} or \eqref{eq:B1onlinehuber}.
\STATE Update $(\mathbf{B}_2^{t+1},\mathbf{B}_3^{t+1})$ from \eqref{eq:B2online} and \eqref{eq:B3online}, respectively.
\STATE Update $(\mathbf{M}_{12}^{t+1},\mathbf{M}_{13}^{t+1})$ from \eqref{eq:LM12online} and \eqref{eq:LM13online}, respectively.
\ENDFOR
\end{algorithmic}
\end{algorithm}


\section{Experimental Validation}\label{sec:simulations}

\begin{figure}
\vspace*{-0.5em}
\centering
\includegraphics[width=0.33\textwidth]{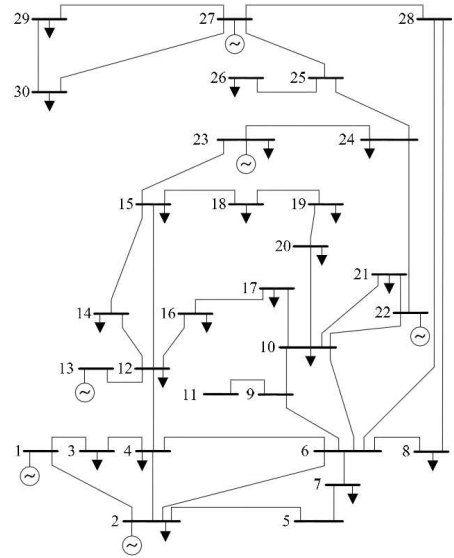}
\vspace*{-0.5em}
\caption{Topology of the IEEE 30-bus system~\cite{PSTCA}.}\label{fig:topology}
\end{figure}

The novel topology recovery approaches are tested next using real load data on the IEEE 30-bus benchmark~\cite{PSTCA}. The latter comprises 18 load buses, 6 generators, and 6 zero-injection buses. The transmission network consists of 41 lines with ratings ranging from 16 to 130~MVA as listed in~\cite{PSTCA}. 

\begin{table}[t]
\renewcommand{\arraystretch}{1.1}
\caption{Generation Offers} \label{tbl:offers} \centering
\begin{tabular}{|c|c|c|c|c|c|}
\hline
\hline
Generator & \multicolumn{5}{c|}{Block offers [MWh,\$/MWh]}\\ 
\hline
1	&	(30,26)	&	(20,36)	&	(20,44)	&	(10,50)	&	\\	
2	&	(20,21)	&	(20,28)	&	(20,35)	&	(20,43)	&	\\
13	&	(15,38)	&	(15,42)	&	(10,47)		&		&\\
22	&	(10,16)	&	(10,27)	&	(10,41)	&	(10,54)	&	(10,66)\\
23	&	(15,34)	&	(15,40)	&		&		&\\
27	&	(30,35)	&	(15,39)	&		&		&\\
\hline
\hline
\end{tabular}
\end{table}

\begin{table}[t]
\renewcommand{\arraystretch}{1.1}
\caption{Average bus degree attained for $(\kappa_1,\kappa_2)$} \label{tbl:sparsity} \centering
\begin{tabular}{|c||c|c|c|c|c|}
\hline
$\kappa_1$~$\backslash$~$\kappa_2$ & 0.001 & 0.01 & 0.1 & 1 & 10\\ 
\hline \hline
0.001	&    0.9	& 1.2	&	1.5	& 	2.5	&	5.9\\ \hline
0.01	&   0.9	& 1.0	& 	1.4	&	2.5	&	5.9\\ \hline
0.1		&    0.8	& 1.0	&	1.6	&	2.4	&	5.9\\ \hline
1		&    0.4	& 2.8	&	1.9	&	2.7	&	5.9\\ \hline
10		&    6.5	& 5.6	&    5.9  &	6.0	&	6.0\\
\hline
\end{tabular}
\end{table}

Regarding offers, the benchmark provides generation capacities and quadratic generation costs~\cite{PSTCA}. To comply with market practices, the generation costs were first approximated by convex piece-wise linear functions yielding the block offers of Table~\ref{tbl:offers}. The original costs were scaled up by 10 to reflect current wholesale electricity cost levels. To model small fluctuations in costs, the nominal offers of Table~\ref{tbl:offers} were shifted by a deviation uniformly distributed in $[-2.5,2.5]$~\$/MWh.  

For consumption, apart from the 18 load buses, generator buses 2 and 23 have load demands too, resulting in a total of 20 loads. The IEEE 30-bus benchmark provides a single realization of load demands. To simulate multiple realistic demands, we used the actual load data publicized for the Global Energy Forecasting (GEF) competition 2012~\cite{kaggle}. These data are the hourly energy consumptions over 20 sites. To match the load levels of the IEEE 30-bus grid, all loads were scaled down by a factor of 7. The 20 demand sequences from December 23, 2007, were assigned to buses so that the average consumption per bus matched the demand specified by the benchmark. Hourly loads were perturbed by a zero-mean Gaussian variation having standard deviation 10 times smaller than the nominal value to account for 5-min load fluctuations.

Real-time prices were generated by solving~\eqref{eq:market} for one day, i.e., 288 5-min intervals, and MCCs were announced separately. Lacking any day-ahead market information, the system was assumed to be dispatched entirely through the real-time market. Out of the 288 dispatches, 3 were infeasible and 45 experienced no congestion (occurred primarily over nighttime). Our experimental validation utilized the remaining $T=240$ MCC price vectors. It is worth stressing that only lines (1,2), (15,23), and (6,28) became congested.

\begin{figure}
\centering
\subfigure[Actual Laplacian matrix.]{
\includegraphics[width=0.4\textwidth]{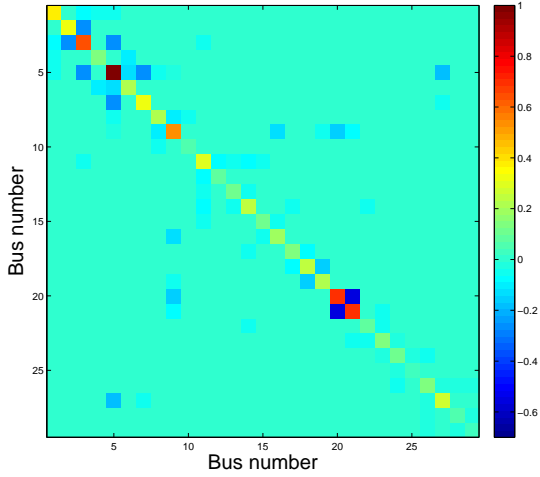}
\label{fig:Laplacian:actual}}
\subfigure[Laplacian matrix found by Alg.~\ref{alg:batch} for $(\kappa_1,\kappa_2)=(1,1)$.]{
\includegraphics[width=0.4\textwidth]{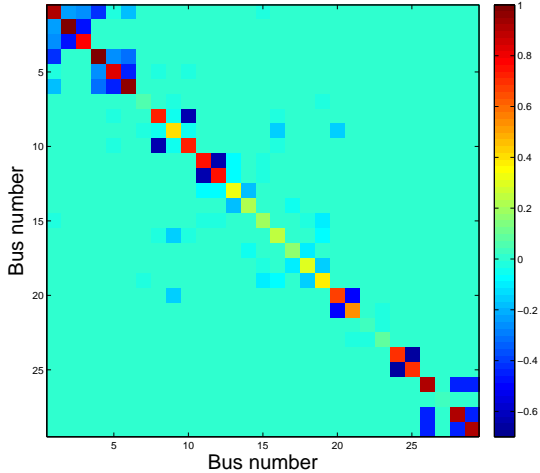}
\label{fig:Laplacian:recovered}}
\caption{Laplacian matrix for the IEEE 30-bus system.}\label{fig:Laplacian}
\end{figure}

Upon collecting prices over an entire day, Alg.~\ref{alg:batch} was run on a 2.4 GHz Intel Core i7 (4GB RAM) laptop computer using MATLAB. Before running the algorithm, parameters $(\kappa_1,\kappa_2)$ were selected. Although such parameters are typically tuned using cross-validation, this methodology becomes cumbersome for our problem. Assuming the average node degree for the grid of interest to be known, $(\kappa_1,\kappa_2)$ were tuned so that the estimate $\hat{\mathbf{B}}$ had the same average degree. Given the scale ambiguity, the algorithm outcome $\hat{\mathbf{B}}$ was normalized by its maximum diagonal entry, and entries with absolute value smaller than $0.01$ were set to zero.

Algorithm~\ref{alg:batch} was run for $(\kappa_1,\kappa_2)$ taking the values $\{10^{-3},10^{-2},10^{-1},1,10\}$. Regarding $\rho$, the convergence rate for the objective (constraint violation) is proportional (inversely proportional) to $\rho$~\cite{oADMM}. For the problem at hand, setting $\rho=10^4$ was empirically observed to provide a good trade-off. Since the average degree of the IEEE 30-bus grid is 2.68, the estimated node degrees obtained in Table~\ref{tbl:sparsity} hint that $(\kappa_1,\kappa_2)$ could be both set to 1. The actual and the recovered Laplacian matrix for the IEEE 30-bus benchmark are shown in Fig.~\ref{fig:Laplacian}.

\begin{figure}
\centering
\includegraphics[width=0.49\textwidth]{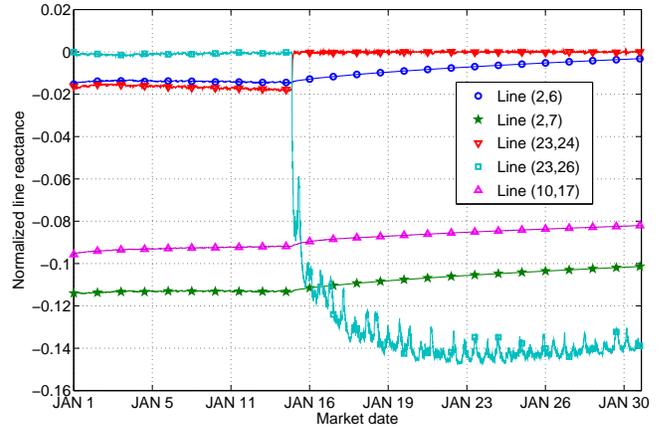}
\vspace*{-1em}
\caption{Tracking lines using streaming pricing data for January 2008.}\label{fig:online}
\vspace*{-1em}
\end{figure}

To evaluate the online scheme, tests on real-time prices collected over January 2008 were conducted. Consumption data were generated by scaling GEF competition loads so that the maximum daily per-site value was 1.6 times the benchmark demands~\cite{kaggle}. Tracking ability was tested by simulating a grid reconfiguration on January 15: lines (2,6) and (23,24) were exchanged for lines (2,7) and (23,26), respectively. Among the 8,928 intervals, infeasible dispatches and dispatches without congestion were ignored yielding 7,220 effective clearings. Alg.~\ref{alg:online} with the update of \eqref{eq:B1onlinehuber} was initialized to the batch solution obtained from Alg.~\ref{alg:batch}. Parameters $\rho$ and $\eta$ were set to $\sqrt{T}$ yielding sublinear regret~\cite{oADMM}, while $\kappa_3$ was set to 1. Figure~\ref{fig:online} depicts the tracking behavior of Alg.~\ref{alg:online}. The estimated normalized reactance for line (10,17), i.e., entry $\hat{\mathbf{B}}_{9,16}$, remained relatively invariant. Line (2,7) was initially erroneously detected as active, yet it was adjusted after Jan. 15, while reactance (2,6) approached zero. Interestingly, the replacement of line (23,24) by (23,26) was promptly detected.

\section{Conclusions}\label{sec:conclusions}
Grid topology recovery using publicly available energy prices was the subject of this work. Upon exploiting the way real-time LMPs are obtained, recovery approaches with complementary strengths were developed. Advances in compressive sampling and online convex optimization proved to be useful for grid topology tracking. Experimental validation using real consumption data on a benchmark grid corroborated the risk of unveiling the power network structure. Numerical tests using a month-long price dataset showed the possibility of tracking grid reconfigurations too. The recovery performance could be enhanced further in envisioned smart grids: In competitive markets, rapidly changing offers and bids could probe the dispatch linear program in a richer way, while market data announced at higher rates could provide even more information. Leveraging heterogeneous market data and characterizing identifiability are interesting research directions.

\appendix

\begin{IEEEproof}[Proof of Proposition~\ref{pro:ell1}]
Strict convexity of $\tfrac{1}{2}\|\mathbf{X}-\mathbf{Y}\|_F^2$ implies that \eqref{eq:Pell1} admits a unique minimizer. First-order optimality conditions assert that $\mathbf{0}$ belongs to the subdifferential of $\|\mathbf{X}\mathbf{z}\|_1 + \tfrac{1}{2}\|\mathbf{X}-\mathbf{Y}\|_F^2$ evaluated at $\hat{\mathbf{X}}$. By definition, the subdifferential of $\|\mathbf{X}\mathbf{z}\|_1$ at $\hat{\mathbf{X}}$ is $\hat{\mathbf{g}}\mathbf{z}'$, where 
\begin{align}\label{eq:module2}
\hat{g}_n':=\left\{\begin{array}{ll}
\sign(\hat{\mathbf{x}}_n'\mathbf{z}) &,~\hat{\mathbf{x}}_n'\mathbf{z}\neq 0\\
s_n:|s_n|\leq 1 &,~\textrm{otherwise}
\end{array}\right.
\end{align}
is the $n$-th entry of $\hat{\mathbf{g}}$, and $\hat{\mathbf{x}}_n'$ denotes the $n$-th row of $\hat{\mathbf{X}}$. Hence, the first-order optimality condition implies that
\begin{equation}\label{eq:module1}
\hat{\mathbf{X}} = \mathbf{Y} - \hat{\mathbf{g}}\mathbf{z}'.
\end{equation}
Unless $\mathbf{z}=\mathbf{0}$ and trivially $\hat{\mathbf{X}}=\mathbf{Y}$, the minimizer $\hat{\mathbf{X}}$ is a rank-one update of $\mathbf{Y}$ granted $\hat{\mathbf{g}}$ is known. To find $\hat{\mathbf{g}}$, post-multiply \eqref{eq:module1} by $\mathbf{z}$ to get $\hat{\mathbf{X}}\mathbf{z} =\mathbf{Y}\mathbf{z} - \hat{\mathbf{g}}z$ whose $n$-th entry reads
\begin{equation}\label{eq:module3}
\hat{\mathbf{x}}_n' \mathbf{z} = \mathbf{y}_n'\mathbf{z}-\hat{g}_n z
\end{equation}
with $z:=\|\mathbf{z}\|_2^2$ and $\mathbf{y}_n$ being the $n$-th row of $\mathbf{Y}$. Given \eqref{eq:module2} and depending on $\mathbf{y}_n'\mathbf{z}$, three cases can be identified for \eqref{eq:module3}: \textbf{(c1)} If $\mathbf{y}_n'\mathbf{z}>z$, then $\hat{g}_n=+1$ and $\hat{\mathbf{x}}_n'\mathbf{z}>0$; \textbf{(c2)} if $\mathbf{y}_n'\mathbf{z}<-z$, then $\hat{g}_n=-1$ and $\hat{\mathbf{x}}_n'\mathbf{z}<0$; and \textbf{(c3)} if $|\mathbf{y}_n'\mathbf{z}|\leq z$, then $\hat{g}_n=\mathbf{y}_n'\mathbf{z}/z$ and $\hat{\mathbf{x}}_n'\mathbf{z}=0$; thus proving the claim.
\end{IEEEproof}

\begin{IEEEproof}[Proof of Proposition~\ref{pro:huber}]
Similar to Prop.~\ref{pro:ell1}, first-order optimality conditions imply that
\begin{equation}\label{eq:huber1}
\hat{\mathbf{X}} = \mathbf{Y} - \alpha\hat{\mathbf{g}}\mathbf{z}'.
\end{equation}
where the $n$-th entry of $\hat{\mathbf{g}}$ is defined as 
\begin{align}\label{eq:huber2}
\hat{g}_n':=\left\{\begin{array}{ll}
\hat{\mathbf{x}}_n'\mathbf{z} &,~|\hat{\mathbf{x}}_n'\mathbf{z}|\leq \kappa\\
\kappa \sign(\hat{\mathbf{x}}_n'\mathbf{z}) &,~\textrm{otherwise}
\end{array}\right.
\end{align}
and $\hat{\mathbf{x}}_n'$ is the $n$-th row of $\hat{\mathbf{X}}$. To find $\hat{\mathbf{g}}$, post-multiply \eqref{eq:huber1} by $\mathbf{z}$ to obtain $\hat{\mathbf{X}}\mathbf{z} =\mathbf{Y}\mathbf{z} - \alpha\hat{\mathbf{g}}z$, whose $n$-th entry reads
\begin{equation}\label{eq:huber3}
\hat{\mathbf{x}}_n' \mathbf{z} = \mathbf{y}_n'\mathbf{z}-\alpha \hat{g}_n z
\end{equation}
with $z:=\|\mathbf{z}\|_2^2$ and $\mathbf{y}_n$ being the $n$-th row of $\mathbf{Y}$. Based on \eqref{eq:huber2}, three cases can be distinguished for \eqref{eq:huber3}: \textbf{(c1)} If $|\mathbf{y}_n'\mathbf{z}|\leq \kappa (1+\alpha z)$, then $\hat{g}_n=\hat{\mathbf{x}}_n'\mathbf{z}$; \textbf{(c2)} if $\mathbf{y}_n'\mathbf{z}>\kappa(1+\alpha z)$, then $\hat{g}_n=\kappa$; and 
\textbf{(c3)} if $\mathbf{y}_n'\mathbf{z}< -\kappa(1+\alpha z)$, then $\hat{g}_n=-\kappa$. Note that for (c1), $\hat{g}_n$ depends on the unknown $\hat{\mathbf{x}}_n$. By substituting $\hat{g}_n$ back into \eqref{eq:huber1} and focusing on the $n$-th row of $\hat{\mathbf{X}}$, we arrive at $(\mathbf{I} + \alpha \mathbf{z}\mathbf{z}')\hat{\mathbf{x}}_n=\mathbf{y}_n$. Invoking the matrix inversion lemma yields $\hat{\mathbf{x}}_n=\mathbf{y}_n- \left(\mathbf{y}_n'\mathbf{z}/(\alpha^{-1}+z)\right)\mathbf{z}$.
\end{IEEEproof}

\bibliographystyle{IEEEtran}
\bibliography{IEEEabrv,myabrv,power}

\end{document}